%% file: DEFT.tex
\documentclass{article}

\usepackage{microtype}
\usepackage{graphicx}
\usepackage{subfigure}
\usepackage{booktabs} 

\usepackage{hyperref}

\usepackage[accepted]{icml2024}
\usepackage{graphicx}
\usepackage{amsmath}
\usepackage{amssymb}
\usepackage{mathtools}
\usepackage{amsthm}

\usepackage[capitalize,noabbrev]{cleveref}

\usepackage{inconsolata}
\usepackage{amsmath}
\usepackage{amsfonts}
\usepackage{amstext}
\usepackage{color}
\usepackage{xcolor}
\usepackage{colortbl,booktabs}
\usepackage{multicol}
\usepackage{multirow}
\usepackage{booktabs}
\usepackage{pifont}
\usepackage{graphicx} 
\usepackage{comment}
\usepackage{color}
\usepackage{xcolor}
\usepackage{algorithm}
\usepackage{algorithmic}
\usepackage{enumitem}
\usepackage{subcaption}
\usepackage[font={small}]{caption}

\usepackage{color}

\usepackage{makecell}
\usepackage{graphicx}
\usepackage{subcaption}

\theoremstyle{plain}

\theoremstyle{definition}

\theoremstyle{remark}

\usepackage[textsize=tiny]{todonotes}

\icmltitlerunning{From PEFT to DEFT: Parameter Efficient Finetuning for
Reducing Activation Density in Transformers}

\begin{document}

\twocolumn[
\icmltitle{From PEFT to DEFT: Parameter Efficient Finetuning for
Reducing Activation Density in Transformers}



\icmlsetsymbol{equal}{*}

\begin{icmlauthorlist}
\icmlauthor{Bharat Runwal}{ind}
\icmlauthor{Tejaswini Pedapati}{ibm}
\icmlauthor{Pin-Yu Chen}{ibm}

\end{icmlauthorlist}

\icmlaffiliation{ind}{Independent Researcher}
\icmlaffiliation{ibm}{IBM Research}

\icmlcorrespondingauthor{Bharat Runwal}{bharatrunwal@gmail.com}
\icmlkeywords{Machine Learning, ICML}

\vskip 0.3in
]



\printAffiliationsAndNotice{} 

\begin{abstract}
Pretrained Language Models (PLMs) have become the de facto starting point for fine-tuning on downstream tasks. However, as model sizes continue to increase, traditional fine-tuning of all the parameters becomes challenging. To address this, parameter-efficient fine-tuning (PEFT) methods have gained popularity as a means to adapt PLMs effectively. In parallel, recent studies have revealed the presence of activation sparsity within the intermediate outputs of the multilayer perceptron (MLP) blocks in transformers. Low activation density enables efficient model inference on sparsity-aware hardware. Building upon this insight, in this work, we propose a novel density loss that encourages higher activation sparsity (equivalently, lower activation density) in the pre-trained models. We demonstrate the effectiveness of our approach by utilizing mainstream PEFT techniques, including QLoRA, LoRA, Adapter, and Prompt/Prefix Tuning, to facilitate efficient model adaptation across diverse downstream tasks. Experiments show that our proposed method, \textbf{DEFT} (Density-Efficient Fine-Tuning), can consistently reduce activation density by up to \textbf{44.94\%} on RoBERTa$_\mathrm{Large}$ and by \textbf{53.19\%} (encoder density) and \textbf{90.60\%} (decoder density) on Flan-T5$_\mathrm{XXL}$ (\textbf{11B}) compared to PEFT, using GLUE and QA (SQuAD) benchmarks respectively. We also introduce \textbf{ADA-DEFT}, an adaptive variant of our DEFT approach, which achieves significant memory and runtime savings during inference. For instance, ADA-DEFT reduces runtime by \textbf{8.79\%}and memory usage by \textbf{17.46\%} in Flan-T5$_\mathrm{XL}$, and by \textbf{2.79\%} and \textbf{2.54\%} respectively in Flan-T5$_\mathrm{XXL}$. Additionally, we showcase that DEFT works complementarily with quantized and pruned models.\footnote{Our Code can be accessed at \href{https://github.com/IBM/DEFT}{https://github.com/IBM/DEFT}  } 
\end{abstract}

\input{text/introduction}
\input{text/related_work}

\input{text/method}

\input{text/experiments}

\input{text/conclusion}

\nocite{langley00}

\bibliography{DEFT}
\bibliographystyle{icml2024}

\newpage
\appendix
\onecolumn


\input{text/appendix}

\end{document}

%% file: text/introduction.tex
\section{Introduction}
\begin{figure*}
    \centering
    \includegraphics[width=\textwidth]{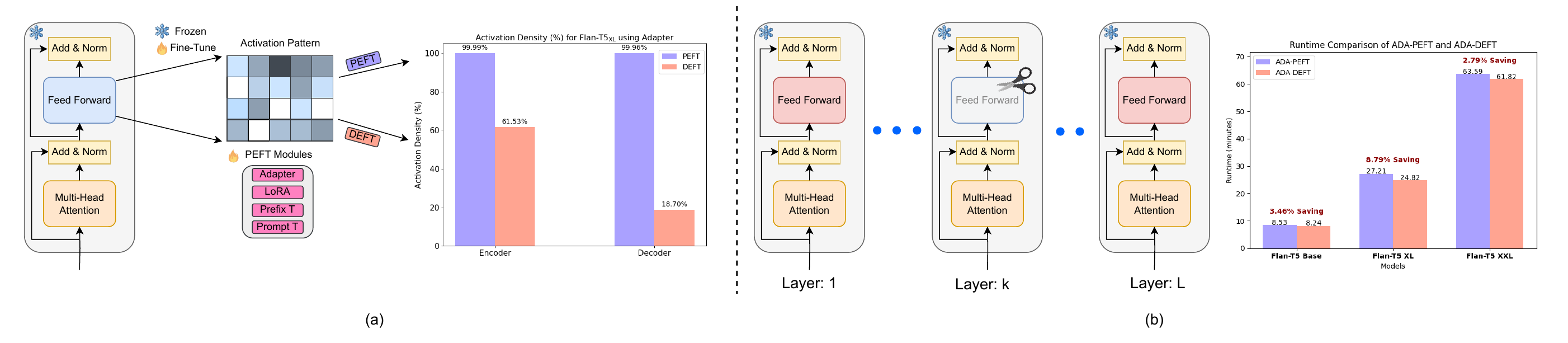}
    \caption{(a) Comparison between the activation density (in the intermediate output of MLP) after adapting to downstream tasks with PEFT and our proposed DEFT method. Both methods use Adapter. (b) ADA-DEFT during inference: based on the learned adaptive layerwise weights, we skip MLP blocks in the ADA-DEFT model, resulting in runtime savings for Flan-T5 models.}
    \label{fig:intro_fig}
\end{figure*}

With the advent of pre-trained large language models (LLMs) \cite{Devlin2019BERTPO,Radford2019_LM_Unsupervised,Raffel2020_T5}, fine-tuning \cite{Howard2018_Universal} these models to adapt to a task has become prevalent. However, training these models and performing inference on them requires a significant amount of time, energy, and memory, thereby resulting in an enormous carbon footprint \cite{Strubell2019_GreenDL}. Some methods to achieve faster and greener inference are by pruning the model parameters \cite{Lee2019_snip,Tanaka2020_Synflow}, pruning the number of heads by analyzing the attention patterns \cite{Behnke2020_lottery_pruning,Voita2019_analyzing_attention,Michel2019_sixteen_heads}, distilling the larger model to a smaller model \cite{Sanh2019_DistilBert}, quantizing the models to convert the weights to a lower precision \cite{Dettmers2022_LLM8bit,Zadeh2020_GOBO_Quant}, using mixture of experts (MOE) \cite{Kudugunta2021_TaskMoE,Rajbhandari2022_DeepSpeedMoE}, etc. 
In contrast, this paper focuses on accelerating the model inference by increasing the activation sparsity in the model. This is achieved by including a penalty for high activation density in the loss function.

Recent studies \cite{Zhang2021MoEficationTF,Li2022_Large_Parsimonious} have shown that in a transformer architecture, specifically in the intermediate outputs of MLP (Multi-Layer Perceptron) blocks with ReLU activations, only a fraction of neurons are activated for a given input, leading to the sparse activation maps as the output. Building upon this observation, we propose a novel density loss that encourages higher activation sparsity in pre-trained models when adapting to downstream tasks, effectively reducing the activation density.

Moreover, the induction of higher activation sparsity holds promising prospects for substantial energy savings, especially on modern hardware acceleration architectures like ASICs (Application Specific Integrated Circuits) \cite{lazzaro2023minimizing}, which leverage zero-skip operations. By promoting sparsity in the activation maps of transformers, hardware can take advantage of zero-skip operations, skipping unnecessary computations on zero-valued activations, resulting in reduced power consumption and more efficient model inference. This energy-efficient approach becomes particularly advantageous for resource-constrained environments or applications with strict energy constraints.

In this work, we present Density-Efficient Fine-Tuning (DEFT) and its variant ADA-DEFT (Adaptive-DEFT), which induces activation sparsity using parameter-efficient fine-tuning (PEFT) techniques. We illustrate DEFT in Figure (\ref{fig:intro_fig}a). The bar plot shows the reduction in Activation Density (\%), which is defined as the number of non-zero values in the intermediate output of MLP layers in transformer blocks averaged over the full validation set. Our proposed DEFT significantly lowers the activation density compared to PEFT. Figure (\ref{fig:intro_fig}b) illustrates our Adaptive DEFT (ADA-DEFT) method, where we skip the MLP block during inference based on the learned adaptive layerwise weights, resulting in runtime savings as shown in the bar plot for Flan-T5 models.

To the best of our knowledge, we are the first to demonstrate that a significant degree of activation sparsity can be attained using a small number of trainable parameters. This is particularly notable in Gaussian Error Linear Unit (GeLU) \cite{Hendrycks2016GaussianEL} models (GeLU and it's variant are the default activation function of state-of-the-art transformer models). Prior studies primarily concentrated on ReLU-based models for investigating activation sparsity \cite{lazzaro2023minimizing}, which are known for their inherent sparsity in activation maps. Our approach, combining PEFT and activation sparsity, paves the way for resource-friendly transformer models across various applications.

%% file: text/related_work.tex
\section{Related Work}

\subsection{Weight Induced Sparsity}
Reducing the number of model parameters results in a model with a lesser memory footprint, reducing the amount of computational resources required to perform the inference.
This can be done by pruning those model parameters whose removal does not deteriorate the model's performance significantly.
To prune the weights at initialization using unstructured pruning, methods such as snip \cite{Lee2019_snip}, grasp \cite{Wang2020_Grasp}, synaptic flow \cite{Tanaka2020_Synflow},  etc can be leveraged. 

Prior arts such as \cite{Li2021_Differentiable_pruning, Voita2019_analyzing_attention, Michel2019_sixteen_heads, Behnke2020_lottery_pruning} sort the heads based on various importance scores and prune the bottom rung. 
For instance, in 
\cite{Michel2019_sixteen_heads}, the importance score is the difference in loss values of when the head is not pruned and when it is pruned. 
Both \cite{Voita2019_analyzing_attention} and \cite{Behnke2020_lottery_pruning} use the confidence of heads as the pruning metric.
Distilling the original model into a smaller model with fewer parameters also achieves the same goal. 
DistilBert \cite{Sanh2019_DistilBert} was obtained by using knowledge distillation on the BERT model during the pre-training phase and is 40\% smaller than BERT while being 97\% as performant as BERT.

\cite{Chen2021_DSEE} used a parameter efficient fine-tuning technique that performs low rank weight updates similar to LoRA.
The user inputs the desired sparsity and whether entire heads must be pruned or any model parameters can be pruned. 
After training, the model parameters and the heads are sorted by the gradient magnitude and are pruned according to the user's preference to achieve the desired sparsity.
In \cite{Sun2023_Wanda}, the importance score for pruning model parameters is calculated as the product of the weight magnitudes and the norms of the input activations. 

\subsection{Activation Induced Sparsity}
Rather than eliminating the parameters apriori, inducing activation sparsity dynamically reduces the latency on a sparsity-aware hardware by reducing the number of computations. \cite{Li2022_Large_Parsimonious} showed that the larger the language models, the sparser their layer outputs are.
Although the output was sparse, there was never a neuron that was never activated.
While 93.5\% of the neurons were activated less than 10\% of the time, the least activated neuron was fired at least 0.001\% of the time. 
Activation sparsity is achieved by thresholding the top-k activation outputs and zeroing out the rest.

\cite{Kurtz2020_inducing_activation} modified ReLU activation of ResNet18 models to fire only if the magnitude of the input is higher than the specified threshold. A special sparsity-aware convolution algorithm is used to accelerate inference. 
As opposed to this, our method naturally induces the activation sparsity during the training owing to our loss function.

%% file: text/method.tex
\section{Methodology} \label{sec:methodology}
\subsection{Background and Notations}
In transformers, the position-wise feed-forward networks employ a two-layer MLP. We measure the activation sparsity at the intermediate output of this two-layer MLP, following the works of \cite{Li2022_Large_Parsimonious} and \cite{Zhang2021MoEficationTF}.

Consider an input \(X \in \mathbb{R}^{B \times K \times d_{\text{model}}}\), where $B$ is the batch size, $K$ is the sequence length and   \(d_{\text{model}}\) denotes the dimensionality of the input features. Given an input matrix \(X\), the output of the two-layer MLP can be described as:
\begin{equation}
Y(X;W_{1},W_{2}) =  f\left(X W_{1}\right) W_{2}
\end{equation}
Here $W_{1} \in \mathbb{R}^{d_{\text{model}} \times d_{\text{ff}}}$ and $W_{2} \in \mathbb{R}^{d_{\text{ff}} \times d_{\text{model}}}$ are the learnable parameters of the MLP layers. $d_{\text{ff}}$ represents the hidden dimension of the MLP block, and $f$ is the non-linear activation function. 

\textbf{Gated-MLP Blocks}: Most large language models (LLMs) currently use the Gated MLP blocks \cite{Shazeer2020GLUVI}. The Gating Mechanism consists of the following computations : 
\begin{equation}
    Y = \left(f(X W_{s}^{T}) \odot (X W_{e}^{T})\right)W_{o}^{T}
\end{equation}
Here $W_{s},W_{e} \in \mathbb{R}^{d_\text{ff}\times d_\text{model}}$ and $W_{o} \in \mathbb{R}^{d_\text{model}\times{d_\text{ff}} }$.

To measure the sparsity of neuron activations, we first define the activation pattern as : 

\begin{align}
O = f\left(X W_{1}\right) \text{~or~} O = f\left(X W_{s}^{T}\right)
\end{align}

The matrix $O \in \mathbb{R}^{B \times K \times d_{\text{ff}}}$ is the activation pattern. Following \cite{Li2022_Large_Parsimonious}, we can define the vector \(s \in \mathbb{R}^{d_{\text{ff}}}\) as the average across the batches and sequence length of matrix $O$ to represents the final feature map. So, we can measure the sparsity of neurons by counting the number of non-zeros in the feature map $s$.

\subsection{Density Loss}
In this section, we introduce our proposed Density loss in DEFT. Our goal is to reduce the activation density (or increase activation sparsity) in MLP blocks for given inputs.

Previous work by  \cite{Li2022_Large_Parsimonious} used a step function to count the number of positive elements precisely, but this operation is non-differentiable and cannot be used for our purpose of reducing activation density in an end-to-end learning setup. Therefore, to approximate the number of non-zero entries in the sparse vector \(s\), we use the hyperbolic tangent function with a scaling parameter \(\beta\) for ReLU-activation based models as defined in \eqref{tanh-loss} for an input feature $x$ with $n$ elements $\{x_i\}_{i=1}^n$. \cite{Krithivasan2020AdversarialSA} used a similar function for their purpose of generating adversarial inputs for sparsity attacks. We also use a differentiable approximation of the $l_0$ norm \cite{lazzaro2023minimizing} for GeLU and other models with activations different from GeLU and ReLU, as defined \eqref{l0-loss} with some hyperparameter $\epsilon>0$. 
\\
\begin{minipage}{0.45\textwidth}
    \begin{equation}\label{tanh-loss}
    \tanh(x,\beta) = \frac{{e^{\beta \cdot x} - e^{-\beta \cdot x}}}{{e^{\beta \cdot x} + e^{-\beta \cdot x}}}
    \end{equation}
\end{minipage}
\hfill
\begin{minipage}{0.45\textwidth}
    \begin{equation}\label{l0-loss}
    \hat{l}_{0} (x,\epsilon) = \sum_{i=1}^{n}\left(\frac{x_{i}^{2}}{x_{i}^{2} + \epsilon}   \right)
    \end{equation}
\end{minipage}

By adjusting the value of \(\beta\) in \eqref{tanh-loss}, we can control the abruptness to approximate the step function (and therefore sparsity) of the values. Higher values of \(\beta\) make the function more closely resemble the step function. In \eqref{l0-loss}, $\epsilon \in \mathbb{R}$ is a parameter dictating the quality of the approximation—lower values of $\epsilon$ correspond to better approximations. In Appendix \ref{approx-non_zero}, we also explored different approximation functions like sigmoid, $l_1$ norm and others. 

We define the density loss \(\mathcal{L}_{\text{density}}(x)\) as follows:
\begin{equation}\label{density_loss}
\mathcal{L}_{\text{density}}(x) = \frac{1}{n}\sum_{l=1}^{L}\sum_{i}g(s_{{l}_{i}})
\end{equation}
Here, \(n\) is the total number of neurons in all the MLP layers, \(L\) is the total number of layers in the transformer, and $s_{l}$ is the feature map after the first dense layer in MLP layer $l$, with $i$ indexing each feature of $s_{l}$. The summation is across all the layers and the elements of the vector \(s_{l}\). The approximation function $g$ can either be the $l_0$ approximation \eqref{l0-loss} or any function in Appendix \ref{approx-non_zero}.

\subsection{Parameter Efficient Fine-tuning (PEFT)}
Fine-tuning a LLM is computationally expensive, as it involves training all the model parameters from scratch.
When adapting a model to multiple datasets, the traditional fine-tuning approach necessitates saving all the trained parameters separately for each dataset, leading to significant storage overhead and computational burden. PEFT addresses this by introducing fewer additional trainable parameters. During fine-tuning, only these parameters are trained, with the rest of the model remaining unchanged. This approach not only reduces computational burden but also minimizes storage demands, as only the new parameters need saving, streamlining the model adaptation process.

Our proposed DEFT is fully compatible with PEFT, and in this paper it adopts Prompt tuning \cite{Lester21_Prompt_tuning}, Prefix tuning \cite{Li2020_Prefix_tuning}, adapters \cite{Houlsby2019_adapter} and Low-rank adaptation (LoRA) \cite{Hu2022_Lora,Dettmers2023QLoRAEF} techniques for demonstration. Each of these techniques is detailed in Appendix \ref{appendix:PEFT}.

\subsection{DEFT: Parameter and Activation Density Efficient Fine-tuning}\label{sec:DEFT}
In our DEFT framework, to efficiently adapt the model without fine-tuning all parameters, we freeze the transformer parameters and only train the aforementioned PEFT modules for downstream tasks. 

For PEFT, we solve the following optimization problem:
\begin{equation}\;\arg\min_{\Phi}\mathcal{L_{\text{T}}}\left(\text{D};\{\Theta,\Phi\}\right) \label{optim-PEFT}
\end{equation}
Here, $\Phi$ represents a set of additional parameters (tunable) in PEFT, while $\Theta$ denotes the set of pre-trained parameters (frozen). The loss function $\mathcal{L_{\text{T}}}$ encapsulates the task-specific objectives and $D$ is the dataset associated with the task.

For our proposed density-efficient fine-tuning (\textbf{DEFT}), i.e. inducing activation sparsity in the MLP layer of transformer blocks, we augment the optimization problem (\ref{optim-PEFT}) by incorporating our density loss (\ref{density_loss}):

\begin{equation}
\begin{aligned}
\arg\min_{\Phi} \mathcal{L}_{\text{total}}=~& \mathcal{L_{\text{T}}}\left(\text{D};\{\Theta,\Phi\}\right) \\
& + \alpha \cdot \mathcal{L_{\text{density}}}\left(\text{D};\{\Theta,\Phi\}\right) \label{optim-DEFT}
\end{aligned}
\end{equation}
Here, the parameter  \(\alpha\) controls the balance between optimizing the performance metric and inducing activation sparsity. Notably, higher values of  \(\alpha\) promote sparser activation maps, although a careful equilibrium is required to ensure minimal impact on performance. We have described the algorithm for DEFT in Algorithm \ref{alg:DEFT}. 

Importantly, the tunable parameters $\Phi$ encompass a versatile range of modules, or compositions thereof, from the choice of \{Adapters, LoRA, QLoRA, Prefix-Tuning, Prompt-Tuning\}, while the original pre-trained parameters $\Theta$ remains frozen. We advocate for these parameter-efficient modules over full-finetuning due to our experimental findings, which reveal that the introduction of a small fraction of trainable parameters (just a few \% of the full model size) suffices to trigger activation sparsity within the MLP blocks. By incorporating these modules, we achieve a twofold efficiency advantage: (1) Facilitating activation sparsity, primed for utilization by hardware accelerators such as ASIC; and (2) Efficient training and storage of these modules, yielding gains in both training time and memory utilization, all while preserving the integrity of downstream task performance.

\begin{algorithm}[tb]
\caption{DEFT}
\label{alg:DEFT}
\begin{algorithmic}[1]
\REQUIRE Dataset $D$, \# of Epochs $E$, Batch Size $B$, \# of Transformer Blocks $L$, Tunable Parameters $\Phi$, Coefficient $\alpha$, Sparsity Approximation Function $g$, Adaptive Sparsity Weights for ADA-DEFT $\{S\}_{i=1}^{L}$

\FOR{$\text{epoch} \gets 1$ to $E$}
    \FOR{$\text{batch} \gets 1$ to $\text{length}(D)$ with batch size $B$}
        \STATE auxiliary variable $\eta \gets []$ 
        \FOR{$\text{i} \gets 1$ to $L$ }
            \STATE $O_i \gets \text{Get the output for } \text{MLP}_i$
            \STATE $\eta.\text{append}(g(O_i))$
            \STATE \textcolor{gray}{$\eta.\text{append}(g(s_i \cdot O_i))$} \hfill \textcolor{gray}{\# ADA-DEFT}
        \ENDFOR
        \STATE Density Loss: $\mathcal{L}_{\text{density}} \gets \text{mean}(\eta)$
        \STATE Total Loss: $\mathcal{L}_{\text{total}} \gets  \mathcal{L}_T + \alpha \cdot \mathcal{L}_{\text{density}}$
        \STATE Update parameters $\Phi$ using $\mathcal{L}_{\text{total}}$ loss
        \STATE \textcolor{gray}{Update parameters $\Phi, S$ using $\mathcal{L}_{\text{total}}$ loss}  \textcolor{gray}{\# ADA-DEFT}
    \ENDFOR
\ENDFOR
\end{algorithmic}
\end{algorithm}

\subsection{ADA-DEFT : Adaptive Parameter and Activation Density-Efficient Fine-Tuning}
In Section \ref{sec:DEFT}, we introduced an additional density loss to induce activation sparsity in pretrained models, with the weight of the loss being a constant $\alpha$, which determines the sparsity in the activations. From prior work in weight pruning \cite{Frankle2018TheLT,Mocanu2017ScalableTO}, it has been observed that performance improves when the sparsity ratio allocation is non-uniform, i.e., each layer is treated differently for the downstream task. Inspired by the non-uniform layerwise weight sparsity, we investigated this non-uniform treatment of activation sparsity for each layer by introducing an extra trainable parameter for each MLP block in the model, formally:

\begin{align}\label{optim-ADA-DEFT}
\arg\min_{\Phi,\text{S}} \mathcal{L}_{\text{total}}=~& \mathcal{L_{\text{T}}}\left(\text{D};\{\Theta,\Phi,\text{S}\}\right) \\
& + \alpha \cdot \mathcal{L_{\text{density}}}\left(\text{D};\{\Theta,\Phi,\text{S}\}\right) \nonumber
\end{align}
Here, $S = \left[S_{1}, S_{2}, \dots, S_{L} \right]$ with $S_{k} \in [0,1]$ for every $k \in [1,L]$ are the trainable parameters for each MLP Block  (adaptive layerwise weights).

In our experiments, we demonstrate that the adaptive layerwise weights help skip some unimportant layers, resulting in memory and runtime savings with minimal impact on downstream performance.

%% file: text/experiments.tex
\section{Experiments}
\paragraph{Datasets: } We evaluated the performance of our method and PEFT techniques using two benchmark datasets: GLUE \cite{wang-etal-2018-glue} and SQuAD \cite{Rajpurkar2016SQuAD1Q}.
GLUE includes various natural language processing tasks. We focused on eight specific datasets: sentiment classification (SST-2), paraphrase detection (MRPC, QQP), natural language inference (MNLI, RTE, QNLI), linguistic acceptability (CoLA), and Semantic Textual Similarity (STS-B).
SQuAD is a well-known reading comprehension benchmark. It comprises of question-answering pairs, requiring the model to provide answers based on given passages. More details about the datasets are in Appendix \ref{appendix:Data_stats}.

\paragraph{Pretrained Language Models:} We used pre-trained RoBERTa$_\mathrm{Large}$ (355M parameters, 24 layers)  \cite{Liu2019RoBERTaAR}; BERT$_\mathrm{BASE}$ (110M parameters; 12 layers)  \cite{Devlin2019BERTPO}; T5$_{\mathrm{SMALL}}$ (60M parameters; 6 encoder and decoder layers), T5$_{\mathrm{BASE}}$ (220M parameters; 12 encoder and decoder layers) \cite{Raffel2019ExploringTL} models; Flan-T5-base (250M parameters, 12 encoder and decoder layers), Flan-T5-xl (3B parameters; 24 encoder and decoder layers), Flan-T5-xxl (11B parameters; 24 encoder and decoder layers) \cite{Chung2022ScalingIL} instruction-tuned models. We also provide additional results with other models, including OPT \cite{Zhang2022OPTOP}, GPT2 \cite{Radford2019LanguageMA}, and ViT \cite{Dosovitskiy2020AnII}) in Appendix \ref{appendix:decoder}. Models \{BERT$_\mathrm{BASE}$, T5$_\mathrm{SMALL}$,T5$_\mathrm{BASE}$, OPT\} uses ReLU activations while the rest of the models utilize GeLU-based activation.

\paragraph{PEFT Modules: } We used \{Adapter, LoRA, Prefix-Tuning (Prefix-T), Prompt Tuning (Prompt-T)\} for \{RoBERTa\} and \{Adapter, LoRA, QLoRA\} for T5 models. These PEFT modules serve as the baselines to be compared with our proposed DEFT method.
More detailed information about the hyperparameters employed in our experiments can be found in Appendix  \ref{appendix:Data_stats}.

\paragraph{Evaluation Metrics: Performance} Regarding the GLUE benchmark, we utilize task-specific evaluation metrics. For the Semantic Textual Similarity (STS-B) dataset, we report the Pearson correlation coefficient. For the CoLA dataset, we use the Matthews correlation coefficient. For MNLI, we report accuracy on the matched validation set, while for all other GLUE tasks, we report accuracy. For the SQuAD dataset, we use the F1 score and Exact-Match score to evaluate performance.

\paragraph{Evaluation Metrics: Efficiency} Beyond task-specific metrics, we evaluate the effectiveness of our method in promoting activation sparsity. We calculate the \textbf{Density} (\%) of activations by identifying the number of non-zero values in the intermediate activation matrices within the MLP block of each transformer layer and averaging these across all layers and the validation set.

We also introduce the \textbf{Density Change} (\%) metric, inspired by the energy consumption ratio concept from \cite{shumailov2021sponge}. This metric compares the sparsity induced by our method to the baseline and is computed as follows:
\begin{equation}\label{density_change}
\resizebox{0.48\textwidth}{!}{$\displaystyle
\text{Density Change (\%)} = \left(\frac{\text{Density}_{\text{PEFT}} - \text{Density}_{\text{DEFT}}}{\text{Density}_{\text{PEFT}}}\right) \times 100
$}
\end{equation}
Here, $\text{Density}{_\text{PEFT}}$ and $\text{Density}{_\text{DEFT}}$, represent the density percentages for baseline PEFT and our DEFT methods, respectively. This formula effectively highlights the reduction in activation density achieved through our approach.

\paragraph{Energy Consumption Ratio} Activation sparsity can be directly leveraged on hardware with zero-skip operations, such as ASIC accelerators. Thus, our DEFT method, which promotes sparser activations, is expected to yield higher energy savings on such hardware. We utilize the energy consumption ratio from \cite{lazzaro2023minimizing}, defined as the ratio between the energy consumed with zero-skipping operations and the energy consumed with standard operations (without zero-skipping). We also report \textbf{Energy Change} (\%), calculated as the relative change between the energy ratios of PEFT and DEFT, normalized by the PEFT energy ratio.

\paragraph{Runtime and Memory Analysis} We report runtime in seconds and memory (in GB) usage for both ADA-DEFT and the baseline ADA-PEFT during evaluation, showcasing practical speedups in inference and reductions in memory storage with ADA-DEFT. This approach provides a direct comparison of real-world performance efficiencies.

\paragraph{Density Loss Hyperparameters: } For our density loss we used $\beta=20$ with tanh approximation Eq. \eqref{tanh-loss} and $\epsilon=1e-07$ with $l_{0}$-approximation Eq. \eqref{l0-loss} and $\alpha=1.0$. We also provide an ablation study for varying these parameters later in Appendix \ref{sec-ablation}. We initialize the adaptive layerwise weights in Eq. \ref{optim-ADA-DEFT} for both the encoder and decoder to 0.80 and then perturbed by adding random noise drawn from a normal distribution with a standard deviation of 0.05.

\subsection{Results on GLUE Benchmark}\label{section-glue}
The performance comparison of different methods on GLUE using the RoBERTa$_\mathrm{Large}$ model is presented in Table \ref{tab:performance_roberta}. This table provides insights into the effects of induced activation sparsity on both performance metrics and activation density in the intermediate layers of the Transformer MLP block with only a few trainable parameters.

\begin{table*}[!t]
\definecolor{Gray}{gray}{0.90}
\newcolumntype{a}{>{\columncolor{Gray}}c}
\centering
\resizebox{1\linewidth}{!}{%
\begin{tabular}{lll|cccccccca}
\Xhline{3\arrayrulewidth}
\textbf{Module (\% Trainable)} & \textbf{Method} &  \textbf{Performance} & \textbf{MNLI} & \textbf{QQP}  & \textbf{QNLI} & \textbf{SST-2} & \textbf{STS-B} & \textbf{MRPC} & \textbf{RTE}  & \textbf{CoLA} & \textbf{Avg.} \\
\Xhline{3\arrayrulewidth} 

\multirow{4}{*} {\textbf{Adapter (1.17\%)}} & 
\multirow{2}{*}{PEFT} & Metric ($\uparrow$)   &  89.83$\pm$ 0.02 & 91.79$\pm$ 0.02 & 94.49 $\pm$ 0.10 & 96.06 $\pm$ 0.44 & 92.31 $\pm$ 0.03 & 89.29 $\pm$ 1.09 & 84.11 $\pm$ 1.47  & 65.43 $\pm$ 0.88 &87.91 \textbf{} \\
& & Density ($\downarrow$) &94.24 $\pm$ 0.02 & 94.06 $\pm$ 0.05  & 94.23 $\pm$ 0.02 & 93.83 $\pm$ 0.12 & 94.41 $\pm$ 0.02 &94.32 $\pm$ 0.08 & 94.54 $\pm$ 0.06 & 94.19 $\pm$ 0.06 & - \\
\cline{2-12} 

& \multirow{3}{*}{DEFT} & Metric ($\uparrow$)  & 89.76 $\pm$ 0.20 & 91.32 $\pm$ 0.07 & 93.67 $\pm$ 0.23 & 96.17 $\pm$ 0.35 &91.76  $\pm$ 0.43 & 89.62 $\pm$ 0.64 & 85.08 $\pm$ 0.17 & 67.14 $\pm$ 1.77 & \textbf{88.06} \\
& & Density ($\downarrow$) & 44.29 $\pm$ 0.55 & 42.38 $\pm$ 0.44 & 46.60 $\pm$ 0.39 & 41.50 $\pm$ 0.53 & 59.85 $\pm$ 2.57 & 63.47 $\pm$ 1.46 & 75.15 $\pm$ 0.41 & 42.01$\pm$ 0.26 & - \\
\cline{3-12} 
& & Density Change (\%) ($\uparrow$)  &53.00 & 54.94 &50.55 & 55.77 &36.61  &32.71 &  20.51 & 55.40& \textbf{44.94 }\\
\Xhline{2\arrayrulewidth} 

\multirow{4}{*} {\textbf{LoRA (1.16\%)}} & 
\multirow{2}{*}{PEFT} & Metric ($\uparrow$)  & 90.53$\pm$ 0.12 & 91.38 $\pm$ 0.04 & 94.71 $\pm$ 0.03 & 95.67 $\pm$ 0.19 & 91.21 $\pm$ 0.24 & 91.63 $\pm$ 0.46 & 81.94 $\pm$ 0.51  & 63.21 $\pm$ 0.01 & \textbf{87.54} \\
& & Density ($\downarrow$) &  94.61 $\pm$ 0.01  & 94.35 $\pm$ 0.15 & 94.49 $\pm$ 0.05 & 93.87 $\pm$ 0.13 & 94.32 $\pm$ 0.03 & 94.28 $\pm$ 0.13 & 94.50 $\pm$ 0.03  & 94.18 $\pm$ 0.06 & - \\
\cline{2-12} 

& \multirow{3}{*}{DEFT} & Metric ($\uparrow$)  & 90.27 $\pm$ 0.21 & 90.79 $\pm$ 0.23 & 93.89 $\pm$ 0.21 & 95.99 $\pm$ 0.33 &  91.52$\pm$ 0.16 & 85.11 $\pm$ 4.81 & 81.40 $\pm$ 1.98  & 61.33 $\pm$ 0.94 & 86.29 \\
& & Density ($\downarrow$) & 43.64 $\pm$ 0.46 & 49.08 $\pm$ 1.29 & 48.65 $\pm$ 0.98 & 45.86 $\pm$ 1.66 & 87.61 $\pm$ 3.35 & 44.00 $\pm$ 5.94 & 85.83 $\pm$ 0.14  & 57.01 $\pm$ 0.95 & - \\
\cline{3-12} 
& & Density Change (\%) ($\uparrow$)  & 53.87 &47.98 &48.51 & 51.15& 7.11 & 53.33 &9.17  & 39.47 & 38.82 \\
\Xhline{2\arrayrulewidth} 

\multirow{4}{*} {\textbf{Prefix-T (1.11\%)}} & 
\multirow{2}{*}{PEFT} & Metric ($\uparrow$)  & 89.99 $\pm$ 0.12 &  89.77 $\pm$ 0.07 & 94.59 $\pm$ 0.05 & 95.64 $\pm$ 0.16 & 90.52 $\pm$ 0.24 &  86.76  $\pm$ 0.91 & 74.84 $\pm$ 1.51 & 59.10 $\pm$0.80 & \textbf{85.15}
\\
& & Density ($\downarrow$) & 94.88 $\pm$ 0.08 & 94.31 $\pm$ 0.21 & 94.14 $\pm$ 0.23 & 94.20$\pm$ 0.24 & 94.24 $\pm$ 0.15 & 93.87 $\pm$ 0.18 & 94.14$\pm$ 0.03  & 93.73 $\pm$ 0.27 & - \\
\cline{2-12} 

& \multirow{3}{*}{DEFT} & Metric ($\uparrow$)   & 89.89 $\pm$0.03 & 89.53 $\pm$ 0.12 & 94.40 $\pm$ 0.06 & 95.72 $\pm$ 0.05 &  90.58 $\pm$ 0.08 & 87.66 $\pm$0.41 & 71.60 $\pm$ 1.51  & 61.35 $\pm$ 1.46 & 85.09 \\
& & Density ($\downarrow$) & 50.83 $\pm$ 0.60 & 46.16$\pm$ 0.27 & 56.19 $\pm$ 0.34 & 51.91 $\pm$ 0.65 & 74.53 $\pm$ 1.48 & 71.88 $\pm$ 0.64 & 76.51 $\pm$ 0.91  & 51.18 $\pm$ 0.97  & - \\
\cline{3-12} 
& & Density Change (\%) ($\uparrow$)  &46.43& 51.06 & 40.31 & 44.89 & 20.91 & 23.43 & 18.73 & 45.4 &36.39 \\
\Xhline{2\arrayrulewidth} 

\multirow{4}{*} {\textbf{Prompt-T (0.31\%)}} & 
\multirow{2}{*}{PEFT} & Metric ($\uparrow$)    & 81.53 $\pm$ 1.93 & 84.17 $\pm$ 0.29 & 80.88 $\pm$ 1.89 & 84.63 $\pm$ 1.68 & 22.66 $\pm$ 14.31 & 71.159 $\pm$ 2.21 & 51.98$\pm$ 3.34  & ${2.18 \pm 3.08}^{*}$ & 59.89 \\
& & Density ($\downarrow$) & 93.78 $\pm$ 0.11 &93.74 $\pm$ 0.15 & 93.10 $\pm$ 0.92 & 93.64 $\pm$ 0.03 &93.77 $\pm$ 0.01 & 93.67 $\pm$ 0.02 & 93.82 $\pm$ 0.02  &  93.55 $\pm$ 0.12 &  - \\
\cline{2-12} 

& \multirow{3}{*}{DEFT} & Metric ($\uparrow$)  & 81.86$\pm$ 1.09 & 83.93 $\pm$ 0.39 & 81.34 $\pm$ 2.35 & 84.51 $\pm$ 1.07 &24.67 $\pm$ 14.11 & 71.07 $\pm$ 2.03 &  51.98 $\pm$ 2.84 & ${0.00}^{*}$  & \textbf{59.92} \\
& & Density ($\downarrow$) & 88.84 $\pm$ 0.41 &89.29 $\pm$ 1.53 & 93.00 $\pm$ 0.10 &92.90 $\pm$ 0.18 & 93.74 $\pm$ 0.01 & 93.65 $\pm$ 0.03 &93.79 $\pm$ 0.02  & 93.05 $\pm$ 0.33 & - \\
\cline{3-12} 
& & Density Change (\%) ($\uparrow$)  &5.27 & 4.75 & 0.11 & 0.79 & 0.03 &0.02  & 0.03   & 0.53 & 1.44 \\
\Xhline{2\arrayrulewidth} 

\end{tabular}
    }
\caption{Performance comparison on GLUE benchmarks with RoBERTa$_\mathrm{large}$. ($^{*}$) denotes unstable training.}
\label{tab:performance_roberta}
\vspace{-2mm}
\end{table*}

Across all datasets, the fine-tuning methods, Adapter, LoRA, Prefix-Tuning (Prefix-T), and Prompt-Tuning (Prompt-T), were evaluated using two approaches: PEFT and DEFT. We observe that all DEFT methods (Adapter, LoRA, Prefix-T, and Prompt-T) generally achieve comparable or better performance to PEFT, with only marginal differences in most cases, while significantly reducing activation density.  The best performance on the GLUE benchmark is observed with DEFT using the Adapter module (\textbf{88.06\%}). The results consistently show that all DEFT methods achieve significantly lower activation density than PEFT.

\paragraph{Reduction in Activation Density: Adapter $>$ LoRA $>$ Prefix-T $>$ Prompt-T.} In all cases, our proposed DEFT method promotes activation sparsity with minimal or no effect on downstream performance. The reductions in activation density range from 0.02\% (Prompt-T, MRPC) to 55.57\% (Adapter, SST-2) across the different datasets and methods. Notably, the method achieving the highest reduction in activation density on the GLUE benchmark is Adapter (44.94\%), followed by LoRA (38.82\%) and Prefix-T (36.39\%). Prompt-T shows the least reduction, at 1.44\%, and we specifically note a training collapse in the CoLA dataset using Prompt-Tuning.

It is important to note that these results are not directly comparable across different modules due to variations in the number of trainable parameters and the locations of additional parameters. For instance, prefix-tuning involves adding trainable parameters to the hidden states. Nonetheless, the generality of DEFT is evident, as all methods effectively promote activation sparsity with minimal impact on downstream performance.

\begin{figure}[t]
\begin{center}
\centerline{\includegraphics[width=\columnwidth]{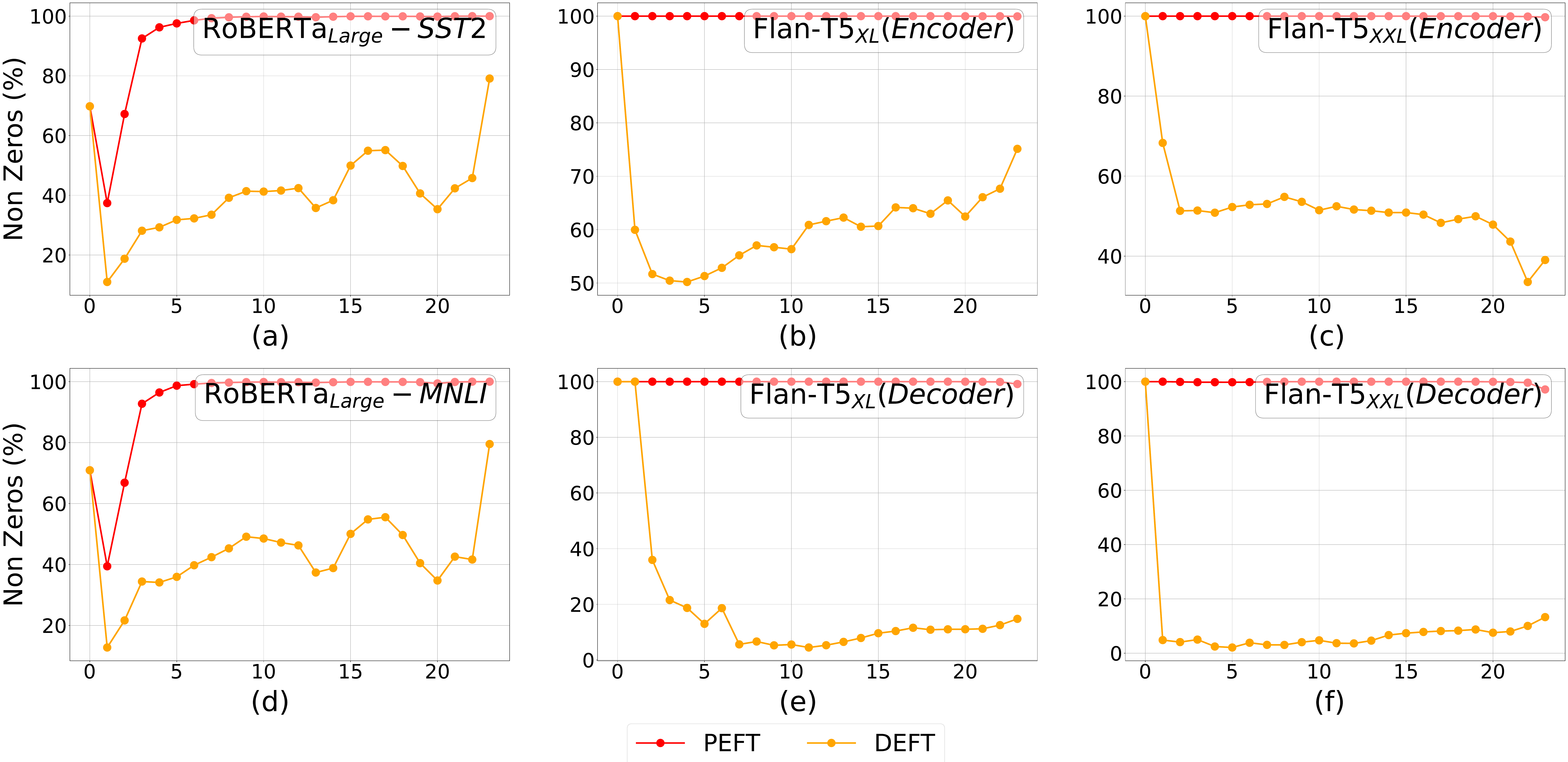}}
\vspace{-2mm}
\caption{\textbf{Percentage of non-zeros (density).} Layerwise non-zeros (\%) for RoBERTa$_\mathrm{Large}$ (a,d), Flan-T5$_\mathrm{xl}$ (b,e) with Adapter, and Flan-T5$_\mathrm{XXL}$ (c.f) with QLoRA on the validation set of different tasks. The x-axis is the layer index.}
\label{fig:layerwise_analyse}
\end{center}
\vspace{-8mm}
\end{figure}

\paragraph{Layerwise Activation Sparsity Analysis.} To delve deeper into the effects of our method, we analyze layerwise activation sparsity in RoBERTa$_\mathrm{Large}$ when paired with Adapter. This analysis is visually represented in Fig. \ref{fig:layerwise_analyse} for the SST-2 (a) and MNLI (d) datasets. Given that RoBERTa$_\mathrm{Large}$ comprises 24 layers, we computed the percentage of non-zero activations across these layers using the validation datasets. The resulting plots reveal a pronounced decrease in non-zero activations at each layer, underscoring DEFT's efficiency in inducing activation sparsity throughout the network's depth.

\paragraph{Energy Consumption Ratio for {$\textbf{RoBERTa}_\mathrm{Large}$. }} In Table \ref{tab:energy_ratio}, we report the energy consumption ratio and energy change (\%). We used SST2 dataset from GLUE benchmark and reported our results on $\textbf{RoBERTa}_\mathrm{Large}$ with different PEFT modules. From the results, we can see that DEFT leads to a reduction in Energy Consumption compared to PEFT on the ASIC simulator. 

\begin{table}[!t]
\small
\centering
\begin{tabular}{cc|c}
\Xhline{3\arrayrulewidth}
 \textbf{Model} &  \textbf{Method} &  \textbf{ER}($\downarrow$)  \\
\Xhline{3\arrayrulewidth} 

\multirow{2}{*}{RoBERTa$_\mathrm{Large}$ (Adapter)} & PEFT  & 0.85 \\
& DEFT  & 0.78 \\
\cmidrule{2-3}
&Energy Change(\%) & \textbf{8.23}\% \\
\cline{1-3} 
\multirow{2}{*}{RoBERTa$_\mathrm{Large}$ (LoRA)} & PEFT  & 0.84\\
& DEFT  & 0.79 \\
\cmidrule{2-3}
&Energy Change(\%) & 5.95\%\\
\cline{1-3} 
\multirow{2}{*}{RoBERTa$_\mathrm{Large}$ (Prefix-T)} & PEFT  & 0.90 \\
& DEFT  & 0.82 \\
\cmidrule{2-3}
&Energy Change(\%) & 8.88\%\\
\cline{1-3} 
\multirow{2}{*}{RoBERTa$_\mathrm{Large}$ (Prompt-T)} & PEFT  & 0.898 \\
& DEFT  & 0.897 \\
\cmidrule{2-3}
&Energy Change(\%) & 0.11\%\\
\cline{1-3} 
\multirow{2}{*}{Flan-T5$_\mathrm{XL}$ (Adapter)} & PEFT  &  1.00
 \\
& DEFT  & 0.87 \\
\cmidrule{2-3}
&Energy Change(\%) & \textbf{13}\%\\
\cline{1-3} 
\multirow{2}{*}{Flan-T5$_\mathrm{XXL}$ (QLoRA)} & PEFT  &  1.00
 \\
& DEFT  & 0.85 \\
\cmidrule{2-3}
&Energy Change(\%) & \textbf{15}\%\\

\Xhline{1\arrayrulewidth} 
\end{tabular}
\caption{Energy Consumption Ratio with different models using ASIC Simulator developed in \cite{shumailov2021sponge}.}
\label{tab:energy_ratio}
\end{table}

\begin{table}[t]
    \centering
    \small    
    \resizebox{\linewidth}{!}{%
    \begin{tabular}{c c c cccc}
        \toprule
        Model&\multirow{2}{*}{Module (\% Trainable)} & \multirow{2}{*}{Loss Type} & \multicolumn{4}{c}{SQuAD} \\
        \cmidrule(lr){4-7} 
        &&& F1($\uparrow$) & Exact-Match($\uparrow$) & Enc-Density($\downarrow$) & Dec-Density($\downarrow$) \\
        \midrule
        \multirow{4}{*}{T5- Small}&\multirow{2}{*}{Adapter (0.33\%)} & PEFT 
        &$82.58\pm0.08$ &$74.48\pm0.07$ &$4.76\pm0.01$&$4.07\pm0.03$ \\
        && DEFT 
        &$82.41\pm0.11$ &$74.19\pm0.13$ &$3.51\pm0.07$&$1.95\pm0.01$  \\
        \cmidrule{3-7}
        && Density Change(\%) & & & 26.26 & 52.08 \\ 
        \cmidrule(lr){2-7}
        (60M)&\multirow{2}{*}{LoRA (0.96\%)} & PEFT 
        &$82.60\pm0.06$ &$74.54\pm0.10$ &$4.80\pm0.01$&$3.97\pm0.01$ \\
        && DEFT
        &$82.38\pm0.09$ &$74.19\pm0.13$ &$3.33\pm0.02$&$1.51\pm0.02$  \\
        \cmidrule{3-7}
        && Density Change(\%) & & & 30.62 & 61.96 \\ 
        \midrule
        \multirow{4}{*}{T5- Base}&\multirow{2}{*}{Adapter (0.40\%)} & PEFT 
        &$88.28\pm0.04$ &$81.19\pm0.05$ &$2.64\pm0.02$&$3.22\pm0.04$ \\
        && DEFT 
        &$88.21\pm0.04$ &$81.08\pm0.12$ &$1.61\pm0.03$&$0.96\pm0.05$  \\
         \cmidrule{3-7}
        && Density Change(\%) & & & 39.01 & 70.19 \\ 
        \cmidrule(lr){2-7}
        (220M)&\multirow{2}{*}{LoRA (0.78\%)} & PEFT
        &$88.33\pm0.03$ &$81.30\pm0.02$ &$2.70\pm0.01$&$3.19\pm0.01$ \\
        && DEFT 
        &$88.42\pm0.04$ &$81.40\pm0.05$ &$1.41\pm0.01$&$0.58\pm0.002$  \\
        \cmidrule{3-7}
        && Density Change(\%) & & & 47.77 & 81.82 \\ 
       \midrule
        \multirow{2}{*}{Flan-T5$_\mathrm{XL}$}&\multirow{2}{*}{Adapter (0.87\%)} & PEFT 
        &$92.81\pm0.03$ &$87.28\pm0.13$ &$99.99\pm0.00$&$99.96\pm0.00$ \\
        && DEFT 
        &$92.52\pm0.03$ &$86.79\pm0.13$ &$61.53\pm0.05$&$18.70\pm0.27$  \\
         \cmidrule{3-7}
        (3B)&& Density Change(\%) & & & 38.46 & 81.29 \\ 
        \midrule
     \multirow{2}{*}{Flan-T5$_\mathrm{XXL}$}&\multirow{2}{*}{QLoRA (1.04\%)} & PEFT 
        &$92.84\pm0.07$ &$86.75\pm0.07$ &$99.97\pm0.00$&$99.82\pm0.01$ \\
        && DEFT 
        &$92.72\pm0.15$ &$87.04\pm0.14$ &$46.80\pm5.65$&$9.38\pm0.41$  \\
         \cmidrule{3-7}
        (11B)&& Density Change(\%) & & & \textbf{53.19}& \textbf{90.60} \\ 
\midrule
       
    \end{tabular}
    }
    \caption{Performance comparison of different methods on Question Answering Dataset (SQuAD) with T5 models.}
    \label{tab:performance_t5}
\end{table}

\subsection{Results on SQuAD Dataset}
Table \ref{tab:performance_t5} offers a comparative analysis of various methods applied to SQuAD using four T5 models. The evaluation focuses on two performance metrics: F1 score and Exact-Match, which measure the accuracy of the model's answers. Additionally, the table provides information on the encoder and decoder activation densities. In our experiments, we introduce the density loss for both encoder and decoder layers with an equal weightage of 1.0, though introducing the density loss for only one of them is also possible. T5$_\mathrm{Small}$ and T5$_\mathrm{Base}$ use ReLU activations, while Flan-T5$_\mathrm{XL}$ and Flan-T5$_\mathrm{XXL}$ use Gated FFN blocks with GeLU-based activation.

\paragraph{Larger models introduce more sparse activation patterns with DEFT.} 
For ReLU-based models (T5$_\mathrm{Small}$ and T5$_\mathrm{Base}$), using the tanh-approximation Eq. \eqref{tanh-loss} in the density loss, DEFT achieves comparable performance to PEFT with notable reductions in activation density. For T5$_\mathrm{Small}$, encoder density reductions range from 26.26\% to 30.62\% and decoder from 52.08\% to 61.96\%. For T5$_\mathrm{BASE}$, encoder reductions are 39.01\% to 47.77\% and decoder 70.19\% to 81.82\%, without sacrificing downstream performance.

For Flan-T5 models, which utilize GeLU activations, we used the $l_{0}$-approximation specified in Eq. \eqref{l0-loss} in the density loss of Eq. \eqref{density_loss}. We investigated our method on larger instruction-tuned models, namely Flan-T5$_\mathrm{XL}$ (3B) and Flan-T5$_\mathrm{XXL}$ (11B). From the results, our method consistently achieves comparable performance with PEFT while significantly reducing activation density for both encoder and decoder layers for larger models. For the Flan-T5$_\mathrm{XL}$ (3B) model with only 0.87\% trainable parameters, we achieve density change (\%) for the encoder \textbf{38.46\%}, while for the decoder, we achieve\textbf{ 81.29\%} when compared to PEFT methods. Similarly, for Flan-T5$_\mathrm{XXL}$ (11B) model, with only 1.04\% trainable parameters, we achieve density change (\%) of \textbf{53.19\%} for the encoder and \textbf{90.60\%} for Decoder compared to PEFT. These findings indicate that DEFT tends to induce sparser activation patterns as the model size increases.

\paragraph{Layerwise Activation Sparsity Analysis.} To provide further insights, we present layerwise non-zero (\%) activation plots in Fig. \ref{fig:layerwise_analyse} using Flan-T5 models on SQuAD. Both models have 24 layers for both encoder and decoder. The plots distinctly show that DEFT significantly reduces the number of non-zero activations in both the encoder (b,c) and decoder (e,f) layers.  A notable observation is the more pronounced reduction in non-zeros in the decoder layers as compared to the encoder layers.

\paragraph{Energy Consumption Ratio for Flan-T5 models.} In Table \ref{tab:energy_ratio}, we report energy consumption ratio and energy change (\%) for Flan-T5 models. We can observe that DEFT leads to a decrease in energy consumption, especially for Flan-T5$_\mathrm{XXL}$, which demonstrates a noteworthy \textbf{15\%} decrease in energy consumption with DEFT compared to PEFT.

\begin{figure}[t]
\begin{center}
\centerline{\includegraphics[width=\columnwidth]{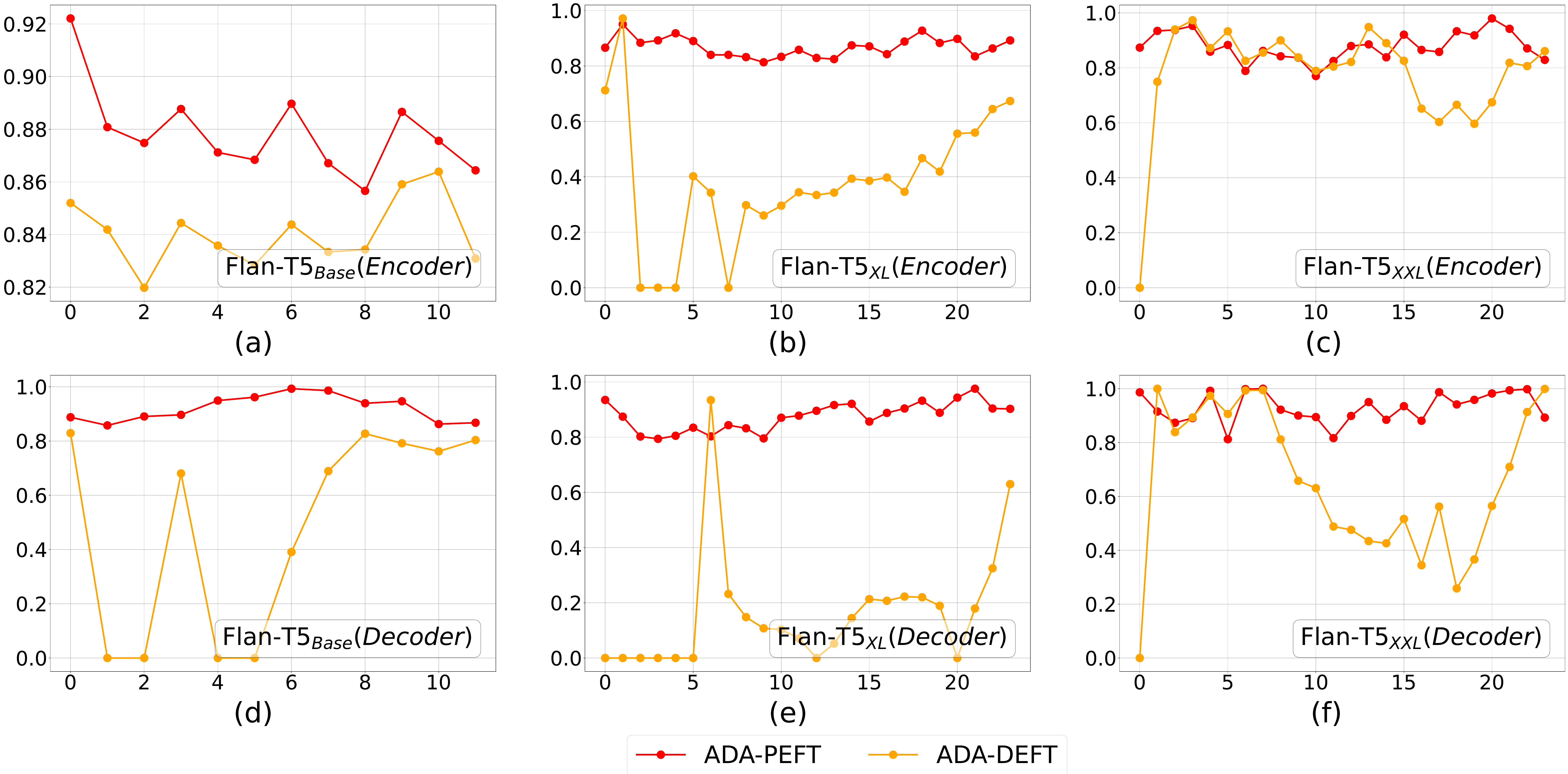}}
\caption{\textbf{Adaptive Layerwise Weights. }Learned Adaptive layerwise weights for ADA-DEFT and ADA-PEFT on SQuAD dataset using Flan-T5 models}
\label{fig:ADA_appendix}
\end{center}
\end{figure}

\paragraph{ADA-DEFT with Flan-T5 models.} We compare the ADA-DEFT method with the baseline ADA-PEFT across various configurations of the Flan-T5 model on the SQuAD dataset in Table \ref{tab:performance_t5_ada_deft}. ADA-DEFT achieves comparable F1 and Exact Match (EM) scores to ADA-PEFT while offering substantial runtime and memory reductions. For the Flan-T5$_\mathrm{BASE}$, ADA-DEFT achieves an F1 score of 89.45 and an EM score of 82.54, with a \textbf{3.46\%} runtime and \textbf{7.37\%} memory savings. For the Flan-T5$_\mathrm{XL}$, it records a \textbf{8.79\%} runtime and \textbf{17.46\%} memory savings, and for the Flan-T5$_\mathrm{XXL}$, the savings are \textbf{2.79\%} in runtime and \textbf{2.54\%} in memory. These efficiencies highlight ADA-DEFT’s capability to reduce resource usage without sacrificing performance. The final learned adaptive layerwise weights for both the encoder and decoder of all Flan-T5 models are shown in Fig. \ref{fig:ADA_appendix}. From the layerwise plots, we observe that for some of the MLP blocks, ADA-DEFT sets the adaptive layer weight to 0, allowing us to skip the MLP block during inference. This results in runtime and memory savings.

\begin{table}[t]
    \centering
    \small    
    \resizebox{\linewidth}{!}{%
    \begin{tabular}{c c c cccc}
        \toprule
        Model&\multirow{2}{*}{Module (\% Trainable)} & \multirow{2}{*}{Loss Type} & \multicolumn{4}{c}{SQuAD} \\
        \cmidrule(lr){4-7} 
        &&& F1 ($\uparrow$) & Exact-Match($\uparrow$) & Runtime (s)($\downarrow$) &Memory (GB)($\downarrow$) \\
       \midrule
        \multirow{2}{*}{Flan-T5$_\mathrm{BASE}$}&\multirow{2}{*}{LoRA (2.67\%)} & ADA-PEFT 
        &$89.60\pm0.02$ &$82.85\pm0.08$ &$511.83\pm3.21$&$0.95\pm0.00$ \\
        && ADA-DEFT 
        &$89.50\pm0.05$ &$82.60\pm0.06$ &$494.13\pm3.38$ &$0.88\pm0.01$\\
         \cmidrule{3-7}
        (250M)&& Saving (\%) & & & $\mathbf{3.46}$& $\mathbf{7.37}$ \\ 
        \midrule
        
        \multirow{2}{*}{Flan-T5$_\mathrm{XL}$}&\multirow{2}{*}{QLoRA (1.99\%)} & ADA-PEFT 
        &$93.11\pm0.04$ &$87.56\pm0.12$ &$1632.89\pm2.99$ &$2.92\pm0.00$\\
        && ADA-DEFT 
        &$92.30\pm0.00$ &$86.56\pm0.06$ &$1489.28\pm3.41$ &$2.41\pm0.03$\\
         \cmidrule{3-7}
        (3B)&& Saving(\%) & & & $\mathbf{8.79}$ &$\mathbf{17.46}$\\ 
        \midrule

     \multirow{2}{*}{Flan-T5$_\mathrm{XXL}$}&\multirow{2}{*}{QLoRA (1.04\%)} &  ADA-PEFT  
        &$92.74\pm0.01$ &$86.50\pm0.13$ &$3815.67\pm3.22$ &$10.61\pm0.00$\\
        && ADA-DEFT 
        &$92.57\pm0.16$ &$86.89\pm0.05$ &$3709.23\pm5.20$&$10.34\pm0.08$ \\
         \cmidrule{3-7}
        (11B)&& Saving(\%) & & & $\mathbf{2.79}$& $\mathbf{2.54}$ \\ 
\midrule
       
    \end{tabular}
    }
    \caption{Performance comparison of \textbf{ADA-DEFT} on Question Answering Dataset (SQuAD) with Flan-T5 models.}
    \label{tab:performance_t5_ada_deft}
\end{table}

In summary, the results show that the Adapter and LoRA modules can maintain competitive performance on SQuAD while achieving notable reductions in activation density and runtime.

\subsection{Pruning of PEFT/DEFT Models}
Here, we explore the effects of model pruning using the WANDA metric \cite{sun2023simple}. The WANDA metric combines weight magnitude and activations to identify parameters for pruning.

Given a weight matrix $W \in \mathbb{R}^{d_{\text{out}}\times d_{\text{in}}}$ and input activations $X \in \mathbb{R}^{N\times{K} \times d_{\text{in}}}$, the importance score of each weight is calculated as:
\begin{equation}
  I_{ij} = |W_{ij}|\cdot||X_{j}||_{2}
\end{equation}
where $||X_{j}||_{2}$ is the $l_2$ norm across the $N\times K$ tokens of the $j$th feature.
\begin{figure}[t]
\begin{center}
\centerline{\includegraphics[width=\columnwidth]{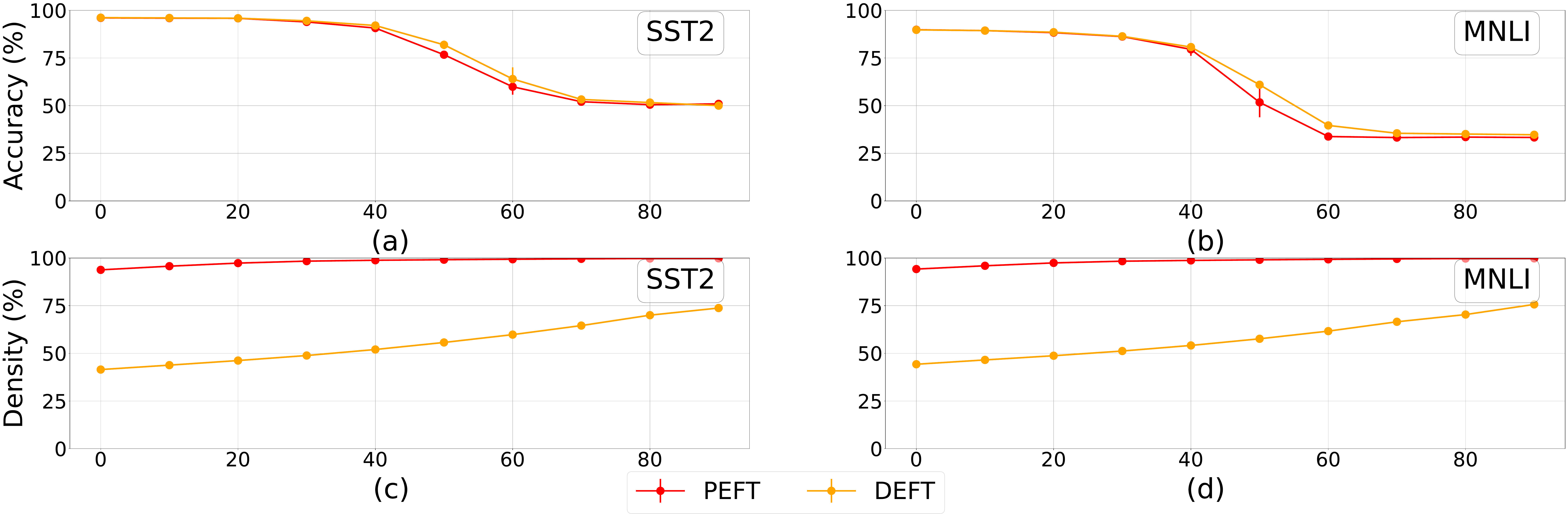}}
\caption{\textbf{Metric v/s Sparsity.} Performance of RoBERTa$_\mathrm{Large}$ with Adapter for different pruning thresholds for MLP block using WANDA metric on the validation set. (a) and (c): Accuracy and Density (\%) on SST2 dataset. (b) and (d):  Accuracy and Density (\%) on MNLI dataset.}
\label{fig:adapt-prune}
\end{center}
\end{figure}

We randomly select 128 samples from the training set of each downstream dataset and use the WANDA metric to prune the first dense layer in the MLP blocks of transformer layers. The results are shown for models first adapted using PEFT/DEFT and then pruned using WANDA.

Fig.\ref{fig:adapt-prune} shows the performance of the pruned RoBERTa$_\mathrm{Large}$ model on the SST-2 and MNLI validation sets. Initially, as sparsity increases, performance remains stable, but beyond a threshold, accuracy declines. DEFT consistently achieves higher or comparable accuracy than PEFT at various sparsity levels while maintaining greater activation sparsity. For example, as shown in Fig. \ref{fig:adapt-prune} (b,d), at 50\% sparsity, the model utilizing DEFT attains an accuracy of 60.96\% with an activation Density of 57.65\%, outperforming PEFT, which achieves only 51.73\% accuracy with a significantly higher Density of 99.09\%. This outcome suggests that DEFT can effectively complement weight pruning methods like WANDA, enabling models to benefit from both activation and weight sparsity.

%% file: text/conclusion.tex
\section{Conclusion}
In this work, we presented our methods DEFT and ADA-DEFT, novel add-on modules to PEFT for inducing activation sparsity in MLP layers of frozen pre-trained transformer blocks. We demonstrate the effectiveness of DEFT for reducing the activation density without hurting downstream performance compared to PEFT by various experiments on GLUE and SQuAD 1.0 benchmark with different models, and with different parameter-efficient modules. Extensive experimental results confirm that our proposed methods provide new means for density-efficient PEFT of pre-trained language models to improve inference efficiency while maintaining similar model performance. We also showcase the effect of pruning with the DEFT models and find that DEFT can be used as a complementary method with weight pruning methods, leading to both activation and weight sparsity. We believe that our proposed DEFT opens a new avenue for density-efficient fine-tuning of pre-trained models.

\section{Acknowledgments}
The authors thank Ming-Hung Chen and I-Hsin Chung at IBM Research for their help and discussion.

%% file: text/appendix.tex
\clearpage

\section{Appendix}
This section provides supplementary information encompassing dataset specifics, code for our experiments and Hyperparameter details for training. We provide more information regarding Parameter-Efficient-Fine-Tuning (PEFT) in Section \ref{appendix:PEFT}; We provide an overview of dataset statistics in Section \ref{appendix:Data_stats}. In section \ref{approx-non_zero}, we delve into supplementary experiments exploring alternative avenues for approximating non-zero activation values. In Section \ref{appendix-glue-bert_base_relu}, we present the experiments on BERT$_\text{BASE}$ model on GLUE benchmark. In Section \ref{sec-ablation}, we provide an ablation study on the hyperparameters used in our DEFT loss. In Section \ref{appendix:decoder}, we present additional experiments with Decoder only models and in Section \ref{appendix:vision_task} on the image classification task.

\subsection{Parameter Efficient Fine-Tuning Methods}
\label{appendix:PEFT}

\paragraph{Prompt tuning} \cite{Lester21_Prompt_tuning}:
Prompt tuning introduces additional trainable parameters called prompt embeddings or soft prompts.
A soft prompt is comprised of  trainable parameters $\Phi$, that are prepended to the embedding of the input tokens for each datapoint.
$\Phi \in  \mathbb{R}^{P \times d_{\text{model}}}$, where P is length of soft prompt and $d_{\text{model}}$  is the dimension of the model's embeddings. 
The soft prompts are initialized randomly and are learnt end-to-end. 
Our input $X$ in (1) would now become $X$ = concat([$\Phi$, emb(input)]).
The soft prompt is the same for all the datapoints in a given downstream task. 
During training,  the entire model's parameters are frozen and only the prompt embeddings are updated during the backpropagation.

\paragraph{Prefix tuning} \cite{Li2020_Prefix_tuning}: In the case of prefix tuning, the trainable prompt embeddings are added to the input at each transformer layer rather than only at the beginning. The prefix tuning embeddings is now $\Phi \in  \mathbb{R}^{L \times P \times d_{\text{model}}}$, where $L$ is the number of transformer layers, P is the length of soft prompt, and $d_{\text{model}}$ is the dimension of the model's embeddings. 

\paragraph{Adapters} \cite{Houlsby2019_adapter}:  The bottleneck adapters consist of a specialized feed-forward layer inserted within each transformer layer.
The adapter architecture typically comprises a down-projection layer, a non-linear activation function, and an up-projection layer.
\begin{equation}
\text{Adapter} = U_{\text{MLP}}(\Omega(D_{\text{MLP}}(\text{inp})))
\end{equation}
where $U_{MLP} \in \mathbb{R}^{I \times m}, D_{MLP} \in \mathbb{R}^{m \times I}$. $I$ is the dimension of the output of the feedback block, $m$ is the dimension of the downward project, and $m \ll I$. 
The specific location of adapters within the transformer block can vary.
In this work, we adopted the placement strategy from \cite{pfeiffer-etal-2020-mad},  where the bottleneck layer is introduced only after each feedforward block in the transformer.

\paragraph{Low-Rank Adaptation (LORA)} \cite{Hu2022_Lora}: During regular fine-tuning, the weights are updated using $W + \Delta W$. In LORA, during the update the model weights $W$ are fixed, $\Delta W$ is decomposed to two lower rank matrices $W_{A}$ and $W_{B}$, and these two matrices are learnt during the training. $W_{A} \in \mathbb{R}^{A \times r}$ and $W_{B} \in \mathbb{R}^{r \times B}$. To save the weights of the model fine-tuned on a new task, only these $W_{A}$ and $W_{B}$ matrices are saved and the final weights are obtained by pre-trained weights $W + W_{A}$$W_{B}$. 

\paragraph{Quantized Low-Rank Adaptation (QLORA)} \cite{Dettmers2023_QLORA}: The memory requirement for all the techniques described so far is comparable to that of fine-tuning. To address this, QLoRa extended LORA where the weights of the model are quantized using double quantization to 4bit NormalFloat.

\subsection{Datasets, Code and Hyperparameters}\label{appendix:Data_stats}
We provide the statistics of the data in table \ref{tab:textdatasplits}. We provide the code for running our experiments in a zip file with the submission.

\paragraph{Pretrained Language Models:} We used pre-trained RoBERTa$_\mathrm{Large}$ (355M parameters, 24 layers)  \cite{Liu2019RoBERTaAR}; BERT$_\mathrm{BASE}$ (110M parameters; 12 layers)  \cite{Devlin2019BERTPO}; T5$_{\mathrm{SMALL}}$ (60M parameters; 6 encoder and decoder layers), T5$_{\mathrm{BASE}}$ (220M parameters; 12 encoder and decoder layers) \cite{Raffel2019ExploringTL} models; Flan-T5-xl (3B parameters; 24 encoder and decoder layers), Flan-T5-xxl (11B parameters; 24 encoder and decoder layers) \cite{Chung2022ScalingIL} instruction-tuned models. We also provide additional results with other models, including OPT \cite{Zhang2022OPTOP}, GPT2 \cite{Radford2019LanguageMA}, and ViT \cite{Dosovitskiy2020AnII}). Models \{BERT$_\mathrm{BASE}$, T5$_\mathrm{SMALL}$,T5$_\mathrm{BASE}$, OPT\} uses ReLU activations while the rest of the models utilize GeLU-based activation.

\paragraph{Training Details:} For training the \{RoBERTa$_\mathrm{Large}$, BERT$_\mathrm{BASE}$\} with adapter and LoRA modules, we used learning rate $lr= 3e-4$, for prefix tuning we used $lr=1e-2$ and for prompt tuning, we used $lr=1e-3$. For all T5 models, we used $lr=3e-5$ for both adapter and LoRA modules. For parameter-efficient fine-tuning on downstream tasks, we freeze the parameters of the pre-trained models and we used 10 epochs with AdamW \cite{Loshchilov2017FixingWD} with Batch size 64. For QLoRA with Flan-T5$_\mathrm{XL}$ and Flan-T5$_\mathrm{XXL}$, we use 4-bit NF4 QLORA with double quantization and paged
optimizer(AdamW-32 bit) \cite{Dettmers2023QLoRAEF} with Batch size of 8 for Flan-T5$_\mathrm{XL}$ and 4 for Flan-T5$_\mathrm{XXL}$. In our experiments, we report all of our results on 3 different random seeds. For ADA-DEFT and ADA-PEFT, we apply our adaptive layerwise weights to the input of the dropout module in the MLP block.  

\paragraph{Evaluation Details:} For evaluating the T5 models on the SQuAD QA generation task, we used a batch size of 64 for all models. The runtime metrics reported for ADA-PEFT and ADA-DEFT were measured using 2 A100 GPUs (80GB) for Flan-T5$_\mathrm{XL}$ and Flan-T5$_\mathrm{XXL}$, and 1 A100 GPU for Flan-T5$_\mathrm{Base}$. The end-to-end inference time is reported in seconds.

\paragraph{Computing Resources:} Our experimental setups for training leveraged 1 RTX8000 with 48GB memory for RoBERTa$_\mathrm{Large}$, BERT$_\mathrm{BASE}$ (ReLU) and T5$_\mathrm{small}$; T5$_\mathrm{base}$ utilized 2 RTX8000 GPUs; Flan-T5$_\mathrm{XL}$ and Flan-T5$_\mathrm{XXL}$ models utilized 4 A100s with 80GB memory, while Flan-T5$_\mathrm{Base}$ used 1 A100 GPU.

\paragraph{Parameter-EFficient Modules:} For the Adapter, we added an adapter block after the feed-forward layer in each transformer block, following \cite{pfeiffer-etal-2020-mad}. Specifically, we used a reduction factor of 16 for the BERT, RoBERTa and Flan-T5 models and 32 for the T5 models. For LoRA we used rank $r=8$, $\alpha = 16$, dropout $=0.10$ for BERT model and rank $r=16$, $\alpha = 32$, dropout $=0.05$ for T5 models and dropout $=0.10$ for Flan-T5 models. For prefix and prompt tuning, we used 60 virtual tokens. We leveraged the PyTorch implementation of Adapter-transformers \cite{pfeiffer-etal-2020-adapterhub} and the PEFT \cite{peft} library, which is built on the Hugging Face Transformers \cite{wolf-etal-2020-transformers} framework. 


\begin{table}[!ht]
    \centering
    \begin{tabular}{|l|l|l|l|}
    \hline
        Dataset & \#Classes & \#Train & \#Validation  \\ \hline \hline
        COLA & 2 & 8550 & 1040  \\ \hline
         MNLI & 3 & 393,000 & 19650 \\ \hline
        MRPC & 2 & 3,670 & 408 \\ \hline
        QNLI & 2 & 105,000 & 5,460  \\ \hline
        QQP & 2 & 364,000 & 40,400  \\ \hline
        RTE & 2 & 2,490 & 277  \\ \hline
        SST2 & 2 & 67,300 & 872 \\ \hline
        STS-B & - & 5,750 & 1,500 \\ \hline
        SQuAD & - & 87,600 & 10,600  \\\hline
    \end{tabular}
        \caption{Number of classes, Number of instances in Train and Validation split in datasets.}
    \label{tab:textdatasplits}
\end{table}

\begin{table*}[!t]
\definecolor{Gray}{gray}{0.90}
\newcolumntype{a}{>{\columncolor{Gray}}c}
\centering
\resizebox{\linewidth}{!}{%
\begin{tabular}{lll|cccccccca}
\hline
\textbf{Module (\% Trainable)} & \textbf{Method} &  \textbf{Performance} & \textbf{MNLI} & \textbf{QQP}  & \textbf{QNLI} & \textbf{SST-2} & \textbf{STS-B} & \textbf{MRPC} & \textbf{RTE}  & \textbf{CoLA} & \textbf{Avg.} \\
\hline 

\multirow{4}{*} {\textbf{Adapter (1.33\%)}} & 
\multirow{2}{*}{PEFT} & Metric ($\uparrow$)   & 83.05 $\pm$ 0.89 & 90.28 $\pm$ 0.15 & 90.84 $\pm$ 0.10 & 92.27 $\pm$ 0.37 & 87.19 $\pm$ 0.34 & 84.26 $\pm$ 0.50 & 56.97 $\pm$ 1.79  & 58.65 $\pm$ 1.94 & 80.44 \\
& & Density ($\downarrow$) & 5.09 $\pm$ 0.05 & 5.28 $\pm$ 0.11 & 5.30 $\pm$ 0.07 & 6.05 $\pm$ 0.15 & 4.95 $\pm$ 0.12 & 5.13 $\pm$ 0.09 & 5.60 $\pm$ 0.35 & 6.09 $\pm$ 0.11 & - \\
\cline{2-12} 

& \multirow{3}{*}{DEFT} & Metric ($\uparrow$)  & 83.68 $\pm$ 0.19 & 89.84 $\pm$ 0.25 & 90.50 $\pm$ 0.11 & 92.36 $\pm$ 0.36 & 87.17 $\pm$ 0.24 & 84.95 $\pm$ 0.59 & 58.19 $\pm$ 1.34 & 58.68 $\pm$ 0.93 &  \textbf{80.67} \\
& & Density ($\downarrow$) & 1.57 $\pm$ 0.02 & 1.17 $\pm$ 0.02 & 1.87 $\pm$ 0.03 & 1.45 $\pm$ 0.06 & 2.91 $\pm$ 0.18 & 2.37 $\pm$ 0.04 & 2.05 $\pm$ 0.13  & 1.10 $\pm$ 0.03 & - \\
\cline{3-12} 
& & Density Change (\%) ($\uparrow$)  & 69.15 & 77.84 & 64.72 & 76.03  & 41.21 & 53.80 & 63.39 & 81.93 & 66.01 \\
\hline

\multirow{4}{*} {\textbf{LoRA (0.27\%)}} & 
\multirow{2}{*}{PEFT} & Metric ($\uparrow$)  &  82.38 $\pm$ 0.46  & 89.09 $\pm$ 0.39 & 90.77 $\pm$ 0.18 & 92.48 $\pm$ 0.38 & 86.76 $\pm$ 0.44 & 86.52 $\pm$ 0.90 & 58.84 $\pm$ 1.00  & 55.86 $\pm$ 0.64 & \textbf{80.34} \\
& & Density ($\downarrow$) &  5.44 $\pm$ 0.11  & 5.39 $\pm$ 0.12 & 5.86 $\pm$ 0.08 & 6.36 $\pm$ 0.21 & 5.57 $\pm$ 0.50 & 5.29 $\pm$ 0.38 & 5.86 $\pm$ 0.17  & 6.51 $\pm$ 0.32 & - \\
\cline{2-12} 

& \multirow{3}{*}{DEFT} & Metric ($\uparrow$)  & 81.83 $\pm$ 0.71 & 88.98 $\pm$ 0.49 & 90.64 $\pm$ 0.13 & 92.34 $\pm$ 0.26 & 86.98 $\pm$ 0.30 & 85.78 $\pm$ 0.73 & 53.94 $\pm$ 1.61  & 56.13 $\pm$ 0.85 & 79.58 \\
& & Density ($\downarrow$) & 1.10 $\pm$ 0.10 & 0.78 $\pm$ 0.03 & 1.52 $\pm$ 0.01 & 1.03 $\pm$ 0.03 & 2.99 $\pm$ 0.19 & 1.70 $\pm$ 0.06 & 0.34 $\pm$ 0.32  & 0.8 $\pm$ 0.03 & - \\
\cline{3-12} 
& & Density Change (\%) ($\uparrow$)  & 79.78 & 85.53 & 74.06 & 83.80 & 46.32 & 67.86 &  94.20 & 87.71  & 77.41 \\
\hline 

\multirow{4}{*} {\textbf{Prefix-T (0.83\%)}} & 
\multirow{2}{*}{PEFT} & Metric ($\uparrow$)  & 81.57 $\pm$ 0.58 &  87.99 $\pm$ 0.93 & 90.32 $\pm$ 0.14 & 92.29 $\pm$ 0.17 & 86.42 $\pm$ 0.40 & 84.66 $\pm$ 1.23 & 59.42 $\pm$ 1.44  & 56.47 $\pm$ 1.72 & \textbf{79.89} \\
& & Density ($\downarrow$) & 5.98 $\pm$ 0.51 & 4.61 $\pm$ 0.27 & 5.21 $\pm$ 0.17 & 5.97 $\pm$ 0.72 & 4.91 $\pm$ 0.46 & 5.45 $\pm$ 0.29 & 6.18 $\pm$ 0.11  & 7.24 $\pm$ 1.32 & - \\
\cline{2-12} 

& \multirow{3}{*}{DEFT} & Metric ($\uparrow$)   & 81.11 $\pm$ 1.45 & 87.99 $\pm$ 0.84 & 89.84 $\pm$ 0.17 & 92.20 $\pm$ 0.26 & 86.62 $\pm$ 0.35 & 84.12 $\pm$ 1.55 & 54.01 $\pm$ 2.26  & 56.18 $\pm$ 1.41 &  79.01 \\
& & Density ($\downarrow$) & 1.05 $\pm$ 0.09 & 0.71 $\pm$ 0.07 & 1.23 $\pm$ 0.04 & 1.02 $\pm$ 0.02 & 1.33 $\pm$ 0.08 & 1.51 $\pm$ 0.01 & 0.77 $\pm$ 0.45  & 0.52 $\pm$ 0.02  & - \\
\cline{3-12} 
& & Density Change (\%) ($\uparrow$)  & 82.11 & 84.60 & 76.39 & 82.91 & 72.91 & 72.29& 87.54 & 92.81 & 81.44 \\
\hline

\multirow{4}{*} {\textbf{Prompt-T (0.03\%)}} & 
\multirow{2}{*}{PEFT} & Metric ($\uparrow$)    & 71.18 $\pm$ 1.14 & 80.15 $\pm$ 0.18 & 80.27 $\pm$ 1.08 & 86.51 $\pm$ 0.34 & 32.66 $\pm$ 8.41 & 70.29 $\pm$ 1.38 & 56.97 $\pm$ 1.64  & 13.68 $\pm$ 14.03 & 61.46 \\
& & Density ($\downarrow$) & 4.39 $\pm$ 0.15         & 4.34 $\pm$ 0.11 & 4.10 $\pm$ 0.14 & 4.88 $\pm$ 0.11 & 4.14 $\pm$ 0.16 & 4.22 $\pm$ 0.05 & 4.40 $\pm$ 0.06  &  3.77 $\pm$ 0.13 &  - \\
\cline{2-12} 

& \multirow{3}{*}{DEFT} & Metric ($\uparrow$)  & 71.01 $\pm$ 1.13 & 80.16 $\pm$ 0.42 & 80.37 $\pm$ 0.79 & 86.40 $\pm$ 0.36 & 32.77 $\pm$ 8.74 & 70.05 $\pm$ 1.34 & 57.26 $\pm$ 2.01 & 13.97 $\pm$ 13.41 & \textbf{61.50} \\
& & Density ($\downarrow$) & 1.91 $\pm$ 0.01 & 1.53 $\pm$ 0.05 & 2.84 $\pm$ 0.24 & 2.18 $\pm$ 0.30 & 4.03 $\pm$ 0.14 & 3.95 $\pm$ 0.12 & 4.19 $\pm$ 0.07  & 2.37 $\pm$ 0.25 & - \\
\cline{3-12} 
& & Density Change (\%) ($\uparrow$)  & 56.49 & 64.75 & 30.73 & 55.33 & 2.66 & 6.40 & 4.77  & 37.13 & 32.28 \\
\hline

\end{tabular}
}
\caption{Performance comparison on GLUE benchmarks with BERT$_\mathrm{BASE}$ (ReLU). }
\label{tab:glue_res}

\vspace{-2mm}
\end{table*}

\begin{table}[!t]
\definecolor{Gray}{gray}{0.90}
\newcolumntype{a}{>{\columncolor{Gray}}c}
\small

\centering
\begin{tabular}{ll|c}
\Xhline{3\arrayrulewidth}
 \textbf{Method} &  \textbf{Performance} &  \textbf{SST2}  \\
\Xhline{3\arrayrulewidth} 

\multirow{2}{*}{PEFT} & Metric ($\uparrow$)  &  92.27 $\pm$ 0.37 \\
& Density ($\downarrow$) &  6.05 $\pm$ 0.15 \\
\cline{1-3} 
\multirow{2}{*}{DEFT ($l_0$)} & Metric ($\uparrow$) & 92.13 $\pm$ 0.30  \\
& Density ($\downarrow$) & 1.41 $\pm$ 0.04\\
\cline{1-3} 
 \multirow{2}{*}{DEFT ($\tanh$)} & Metric ($\uparrow$)  & {92.36 $\pm$ 0.36} \\
 & Density ($\downarrow$) & 1.45 $\pm$ 0.06\\
\cline{1-3} 
 \multirow{2}{*}{DEFT ($\text{sigmoid}$)} & Metric ($\uparrow$)  & {92.02 $\pm$ 0.05} \\
 & Density ($\downarrow$) & 1.51 $\pm$ 0.01\\
\cline{1-3} 
 \multirow{2}{*}{DEFT ($l_1$)} & Metric ($\uparrow$)  & {50.91 $\pm$ 0.00} \\
 & Density ($\downarrow$) & 0.50 $\pm$ 0.00\\
\Xhline{1\arrayrulewidth} 

\end{tabular}
\caption{Performance comparison with different approximations using BERT$_\mathrm{BASE}$ (ReLU) with Adapter module (1.33\% trainable parameters). }
\label{tab:diff_approx}
\end{table}


\subsubsection{Approximation of Non-Zero Activations}\label{approx-non_zero}
\paragraph{$\boldsymbol{\tanh} > \boldsymbol{l_0} > \text{sigmoid}> \boldsymbol{l_1}$.} We compared DEFT with different approximations in Table \ref{tab:diff_approx}. We found that among the four of them; the $\tanh$ approximation works best for the ReLU-based model. Other approximation sigmoid and $l_0$ also performed well, however $l_1$ norm fails to retain good performance while reducing activation density, highlighting that a naive approximation without proper hyperparameter search can lead to unstable training.

Our choice of using \(\tanh\) for the ReLU model is motivated by the results above. Since ReLU sets negative values to zero, only the part of \(\tanh(x)\) where \(x \geq 0\) is relevant for our case. However, we can't use \(\tanh(x)\) for models with GeLU and other activation function variants because these allow negative values. Therefore, for all other models utilizing GeLU-based activation functions, we employed the \(l_0\) approximation method.

\subsubsection{Results on GLUE Benchmark with BERT$_{\text{BASE}}$ (ReLU)}\label{appendix-glue-bert_base_relu}
We provide the results on GLUE becnhmark in Table \ref{tab:glue_res} on BERT$_{\text{BASE}}$ (ReLU) model with different PEFT modules. We observe that all DEFT methods, including Adapter, LoRA, Prefix-T, and Prompt-T, generally achieve comparable performance to PEFT, with only marginal differences in most cases while significantly reducing the activation density. The results reveal that all methods consistently achieve significantly lower activation density compared to PEFT. We observe in all the cases, our proposed method DEFT promotes activation sparsity with minimal or no effect on the downstream performance. The reductions in activation density range from 2.66\% (Prompt-T, STS-B) to 94.20\% (LoRA, RTE) across the different datasets and methods. Notably, the method with the highest reduction in activation density on GLUE benchmark is Prefix-T (81.44\%), followed by LoRA  (77.41\%) and Adapter  (66.01\%). Prompt-T achieves the lowest reduction among the methods (32.28\%).

\begin{table}[!ht]
    \centering

    \begin{tabular}{|c|c|c|}
    \hline
        Models  &Batch Size&lr  \\ \hline
        ViT$_{\text{Base}}$ & 64 & 2e-5   \\ \hline
        ViT$_{\text{Large}}$ & 32 & 2e-5  \\ \hline
        GPT2 & 32 & 3e-4 \\ \hline
        OPT & 32 & 3e-4     \\ \hline
    \end{tabular}
        \caption{Hyperparameters of different models used for additional study}
    \label{tab:hyper_add}
\end{table}

\subsubsection{Ablation Study}\label{sec-ablation}
In this section, we perform an ablation study on two parameters used in our proposed method DEFT in Eq. (\ref{optim-DEFT}): $\alpha$ (weight for the density loss) and $\epsilon$ (used in the $l_{0}$ approximation in Eq. \eqref{l0-loss}). For this study, we used the RoBERTa$_\mathrm{Large}$ model with SST2 dataset.

\begin{figure}[t]
\begin{center}
\centerline{\includegraphics[width=\columnwidth]{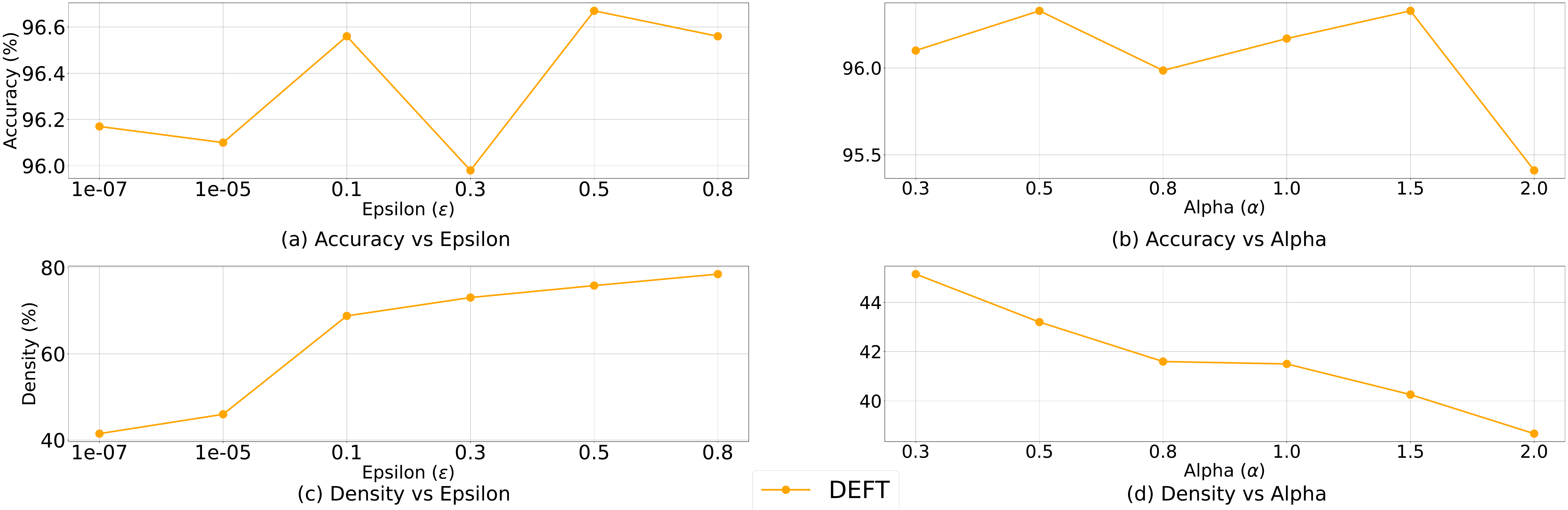}}
\caption{\textbf{Ablation Study.} Varying $\alpha$ and $\epsilon$ parameters using Adapter module with RoBERTa$_\mathrm{Large}$ on SST2 dataset.}
\label{fig:ablation-roberta}
\end{center}
\end{figure}

First, we investigate the impact of varying $\alpha$ on the accuracy and density. Fig. \ref{fig:ablation-roberta} presents the results, showing the accuracy (b) and density (d) for different values of $\alpha$, while keeping $\epsilon$ fixed at $1e-07$ on the SST2 dataset. As we increase the weightage, we observe a decrease in density as the activations in the intermediate layer of the MLP become sparser but the impact on accuracy is minimal degradation. Hence, there is a trade-off between sparsity and downstream performance, analogous to the layerwise sparsity-ratio (\%) in weight sparsity.

Next, we examine the impact of varying $\epsilon$ on the accuracy and density. Fig. \ref{fig:ablation-roberta} illustrates the results, displaying the accuracy (a) and density (c) for different values of $\epsilon$, with $\alpha$ fixed at 1.0 on the SST2 dataset. As we increase $\epsilon$, we observe a rapid increase in density initially, which then saturates. Notably, our method's resilience remains evident as fluctuations in the $\epsilon$ parameter exhibit a minimal impact on the overall performance (accuracy).

\begin{figure}[t]
\begin{center}
\centerline{\includegraphics[width=\columnwidth]{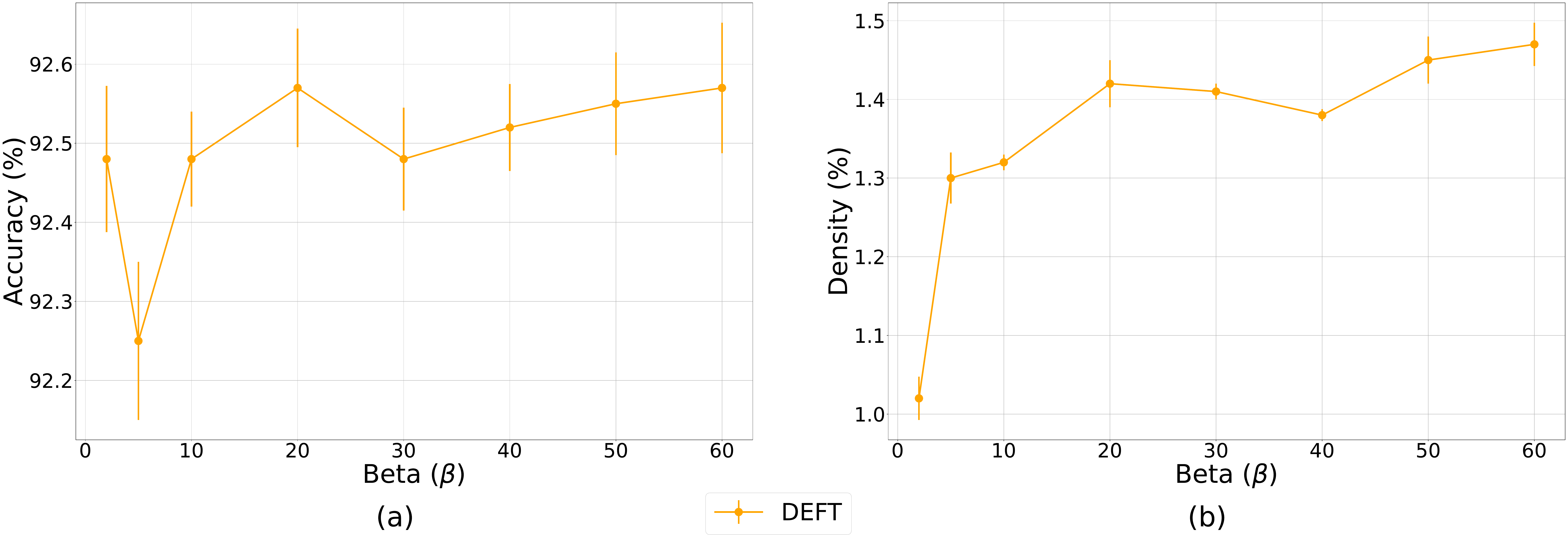}}
\caption{\textbf{Ablation Study.} Varying $\beta$ using Adapter module with BERT$_\mathrm{Base}$ on SST2 dataset.}
\label{fig:ablation-bert}
\end{center}
\end{figure}

Lastly, In Fig.\ref{fig:ablation-bert}, we investigate the impact of varying $\beta$ in Eq. (\ref{tanh-loss}), we can see that as we increase $\beta$, the density starts increasing but accuracy remains nearly the same, which shows that our method is not much sensitive to $\beta$.

\subsubsection{Experiments with Decoder only models} \label{appendix:decoder}
We additionally tested our method with Decoder only models \{OPT, GPT2\} in Table \ref{appendix:extras}. We used SST-2 dataset from GLUE. For GPT2, we used an adapter with reduction factor 8 for our experiments and for OPT models, we used LoRA with rank $r=8$.

From the results, we can see that even for Decoder-only models, our method leads to a significant reduction in activation density both for OPT and GPT2 models. 

\subsubsection{Experiments on Image Classification task} \label{appendix:vision_task}
We tested the image classification task with ViT model \cite{Dosovitskiy2020AnII} on CIFAR-10 \cite{krizhevsky2009learning} dataset. For ViT, we used an adapter with a reduction factor of 8 for our experiments. For the CIFAR-10, we report the accuracy on the Test set containing 10k images. 

From the results in Table \ref{appendix:extras}, we can see that DEFT leads to reduction in activation density; for example, in ViT$_{\text{Large}}$, we see DEFT achieves activation density of 70.33\% while PEFT's density is quite high 99.91\%.   


\begin{table}[!ht]

\centering
\resizebox{\linewidth}{!}{%
\begin{tabular}{ll|ccccc}
\hline
\textbf{Method} &  \textbf{Performance} & \textbf{ViT$_{\text{Base}}$ (86M)} & \textbf{ViT$_{\text{Large}}$ (307M)}  & \textbf{OPT (125M)} & \textbf{OPT (350M)} & \textbf{GPT2 (117M)}  \\

\multirow{3}{*}{PEFT} & Trainable (\%)   &  (2.04\%) & (2.04\%) & (0.24\%)& (0.24\%) & (1.87\%) \\

\hline 
& Metric ($\uparrow$)  & 98.13 $\pm$ 0.03 & 99.04 $\pm$ 0.06 & 91.50$\pm$0.11 & 93.34$\pm$0.46  & 88.38 $\pm$ 0.19  \\
 & Density ($\downarrow$) & 84.84 $\pm$ 0.09 & 99.91 $\pm$ 0.00 &  7.80$\pm$0.26 & 8.45$\pm$0.14 & 99.96 $\pm$ 0.00  \\

\hline
 \multirow{3}{*}{DEFT} & Metric ($\uparrow$)  & 97.75 $\pm$ 0.09 & 98.31 $\pm$ 0.08 &  91.86$\pm$0.11 & 92.15$\pm$0.51 & 88.34 $\pm$ 0.39 \\
 & Density ($\downarrow$) & 77.74 $\pm$ 0.09 & 70.33 $\pm$ 2.47 &  1.24$\pm$0.002 & 1.87$\pm$0.01 &70.51 $\pm$1.15  \\
\hline 
 & Density Change (\%) ($\uparrow$)  & 8.38 & 29.61 &  84.10 & 77.87 & 29.46\\

\hline

\end{tabular}
}
\caption{Peformance Comparison on different models with PEFT and DEFT. }
\label{appendix:extras}
\end{table}